\documentclass{article}
\usepackage{ijcai26}

\usepackage{times}
\usepackage{soul}
\usepackage{url}
\usepackage[hidelinks]{hyperref}
\usepackage[utf8]{inputenc}
\usepackage[small]{caption}
\usepackage{graphicx}
\usepackage{amsmath}
\usepackage{amsthm}
\usepackage{booktabs}
\usepackage{algorithm}
\usepackage{algorithmic}
\usepackage[switch]{lineno}
\usepackage[table]{xcolor}
\usepackage{todonotes}
\usepackage{pifont} 
\newcommand{\cmark}{\ding{51}} 
\newcommand{\xmark}{\ding{55}} 
\definecolor{lila}{RGB}{230,220,245}
\newcommand{\citet}[1]{\citeauthor{#1}~[\citeyear{#1}]}

\title{SynCABEL: Synthetic Contextualized Augmentation for Biomedical Entity Linking}

\author{
Adam Remaki$^1$
\and
Christel Gérardin$^{1,2}$\and
Eulàlia Farré-Maduell$^3$\and
Martin Krallinger$^3$\And \\
Xavier Tannier$^1$\\
\affiliations
$^1$ Sorbonne Université, Inserm, Université Sorbonne Paris Nord, Limics, 75006 Paris, France\\
$^2$ Service de médecine interne, Hôpital Tenon, Assistance Publique - Hôpitaux de Paris, Paris, France\\
$^3$ Barcelona Supercomputing Center, Barcelona, Spain\\
\emails
adam.remaki@etu.sorbonne-universite.fr, 
christel.gerardin@aphp.fr, 
\{eulalia.farre,martin.krallinger\}@bsc.es, 
xavier.tannier@sorbonne-universite.fr
}

\begin{document}

    \maketitle
\begin{abstract}
We present \textbf{SynCABEL} (Synthetic Contextualized Augmentation for Biomedical Entity Linking), a framework that addresses a central bottleneck in supervised biomedical entity linking (BEL): the scarcity of expert-annotated training data. SynCABEL leverages large language models to generate context-rich synthetic training examples for all candidate concepts in a target knowledge base, providing broad supervision without manual annotation. We demonstrate that SynCABEL, when combined with decoder-only models and guided inference, establishes new state-of-the-art results across three widely used multilingual benchmarks: MedMentions for English, QUAERO for French, and SPACCC for Spanish. Evaluating data efficiency, we show that SynCABEL reaches the performance of full human supervision using up to 60\% less annotated data, substantially reducing reliance on labor-intensive and costly expert labeling. Finally, acknowledging that standard evaluation based on exact code matching often underestimates clinically valid predictions due to ontology redundancy, we introduce an \mbox{LLM-as-a-judge} protocol. This analysis reveals that SynCABEL significantly improves the rate of clinically valid predictions. Our synthetic datasets, models, and code are released to support reproducibility and future research:
    \begin{itemize}
        \item \href{https://hf.co/collections/Aremaki/syncabel}{\textbf{HuggingFace Datasets \& Models}}
        \item \href{https://github.com/Aremaki/SynCABEL}{\textbf{GitHub Repository}}
    \end{itemize}
\end{abstract}

\section{Introduction}

Biomedical Entity Linking (BEL) is the task of mapping spans of text in biomedical documents to concepts in knowledge bases (KBs) such as the UMLS or SNOMED CT. 

\begin{figure}[t]
    \centering
    \includegraphics[width=\linewidth]{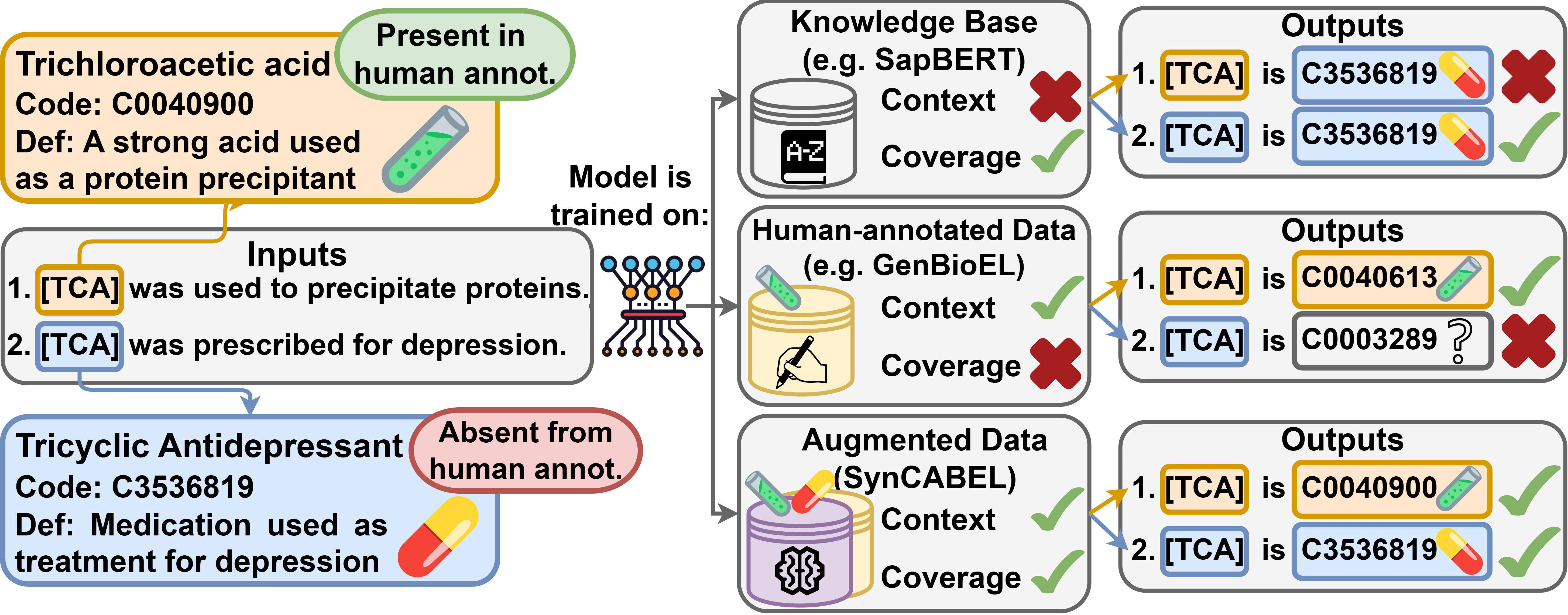}
    \caption{Illustration of biomedical entity linking annotation scarcity: a context-free model fails without context, a supervised context-aware model trained on human annotations fails on unseen concepts, while a SynCABEL-augmented model leverages synthetic data to recover the correct concepts.}
    \label{fig:intro}
\end{figure}

Early approaches relied on context-free lexical matching or self-supervised models trained from the KB alone. However, they struggle with ambiguity: terms such as “discharge” (release vs. secretion) or abbreviations such as “AS” (Aortic Stenosis vs. Ankylosing Spondylitis) may refer to distinct concepts depending on context~\cite{newman-griffis_ambiguity_2021,xu2007study}. These limitations motivated context-aware BEL models, which encode mentions with surrounding text to disambiguate surface forms. Although effective when sufficiently trained, they introduce a major constraint: the need for large-scale, high-quality annotated datasets, which are scarce in the biomedical domain. Creating such resources is labor-intensive, requiring clinical experts to match mentions to ontology concepts. This scarcity limits generalization and creates a bottleneck for supervised BEL, as illustrated in Figure~\ref{fig:intro}.

To address this annotation scarcity issue, we introduce \textbf{SynCABEL} (\textbf{Syn}thetic \textbf{C}ontextualized \textbf{A}ugmentation for \textbf{B}iomedical \textbf{E}ntity \textbf{L}inking), a novel framework designed to enhance context-aware BEL by leveraging large language models (LLMs) to generate context-rich training instances for all candidate concepts in a KB. Our contributions are listed as follows:

\begin{itemize}
    \item We release the first large-scale multilingual synthetic dataset for BEL in English, French, and Spanish.
    \item We show that SynCABEL matches full human-annotated training performance with substantially fewer human annotations.
    \item We demonstrate that combining recent decoder-only models with SynCABEL and guided inference achieves state-of-the-art performance on multiple BEL benchmarks.
\end{itemize}

Additionally, we developed an LLM-as-a-judge evaluation protocol for BEL that goes beyond exact code matching by assessing the semantic relationship between predicted and gold concepts, including equivalence, broader, narrower, and unrelated relations.

\section{Related Work}

\subsection{BEL Systems}

Research in BEL has evolved through several distinct paradigms, each addressing the problem with different strengths and limitations~\cite{chen_comprehensive_2026}.

\paragraph{Rule-based Systems.} Early systems like cTAKES~\cite{savova_mayo_2010}, and SciSpacy~\cite{neumann_scispacy_2019} relied on heuristic string matching. While widely adopted, their reliance on static dictionaries limits generalization to predefined lexical variants, causing failures when a mention refers to an existing concept using a surface form not listed among its synonyms in the ontology.

\paragraph{Context-free Bi-encoder Models.} Transformer-based models like SapBERT~\cite{liu_self-alignment_2021,liu_learning_2021} and CODER~\cite{yuan_coder_2022} use self-supervised contrastive learning on synonyms. While efficient for embedding names in a shared space without annotated data, they ignore surrounding context, limiting their ability to resolve ambiguity.

\paragraph{Prompting LLMs.} LLMs have been used via prompting to simplify mentions before linking~\cite{borchert_improving_2024,vollmers_contextual_2025}, as re-rankers~\cite{ye_llm_2025}, or guided via constrained decoding without training~\cite{lin_guiding_2025}. However, prompting-based methods often yield inferior results compared to specialized EL models.

\paragraph{Contextualized Bi-encoder Models.} These systems encode mentions with context and candidates in a shared space. Unlike context-free models, they explicitly learn to capture contextual information. ArboEL~\cite{agarwal_entity_2022}, for example, adapts BLINK~\cite{wu_scalable_2020} and introduces a graphical arborescence objective to model cross-document coreference through directed spanning trees, achieving state-of-the-art results on MedMentions~\cite{mohan_medmentions_2018}.

\paragraph{Generative Models.} These models treat entity linking as conditional text generation, directly producing concept identifiers. GENRE~\cite{cao_autoregressive_2021} introduced constrained decoding to ensure valid outputs and strictly improve memory efficiency by avoiding dense indexes. Currently, biomedical implementations like GenBioEL~\cite{yuan_generative_2022} are built on pretrained encoder-decoder architectures. The general-domain InsGenEL~\cite{xiao_instructed_2023} has extended this approach by fine-tuning a decoder-only model, though it is optimized for a joint mention detection and linking objective and not for entity linking only.

\subsection{BEL Data Augmentation}
Despite architectural differences, both retrieval and generative methods depend on annotated datasets that currently cover only a fraction of the KB and lack variability. This scarcity creates a bottleneck for developing robust, generalizable BEL systems, motivating alternative data creation strategies to improve scalability and coverage while reducing annotation costs.

\paragraph{Weak and Distant Supervision.} Approaches using exact string matching in large corpora (e.g., PubMed, Wikipedia) to create training data~\cite{zhang_knowledge-rich_2022,vashishth_improving_2021,wang2023novel} are scalable but noisy due to synonym ambiguity and non-clinical contexts.

\paragraph{Template-Based Synthetic Data.} To better capture entity semantics, \citet{yuan_generative_2022} used UMLS definitions as templates, filled with synonyms. This provides clean, concept-focused data but lacks the lexical and contextual diversity of natural language.

\paragraph{LLM-generated Synthetic Data.} \citet{xin_llmael_2025} and \citet{chen_enhancing_2025} have demonstrated the effectiveness of LLMs for generating additional training examples, but restrict synthesis to concepts already observed during training. While this increases contextual diversity and improves generalization across known entities, it does not address annotation scarcity for entirely unseen concepts. Relatedly, \citet{josifoski_exploiting_2023} leverage synthetic data to mitigate data scarcity, but their approach targets a different task setting, namely relational triple extraction.

\subsection{Positioning of SynCABEL}

SynCABEL advances BEL on two fronts. First, it introduces a data augmentation strategy that generates examples for the entire space of candidate concepts in the KB. It overcomes the limitations of prior work that either used templates for full coverage (low quality) or LLMs only for training-set concepts (limited coverage). Second, it is the first approach to fine-tune a decoder-only model specifically for BEL, allowing us to leverage recent improvements in large foundation models.

\section{Methodology}

\subsection{Problem Statement}
Let $\mathcal{E}$ represent the set of candidate concepts from a given KB. Each concept $e \in \mathcal{E}$ is associated with a semantic group $g_e$ (e.g., \textit{Disorders}) and multiple synonyms $S_e = \{s_e^1, \ldots, s_e^{n_e}\}$. Given a mention $m$ with left context $c_l$, right context $c_r$ and semantic group $g_e$, our objective is to predict the correct concept $e_m$ that the mention $m$ refers to.

\subsection{Generative BEL Training Objective}
We adopt the autoregressive formulation from \citet{cao_autoregressive_2021}. The input sequence $x$ is defined as:
\begin{equation*}
x = c_l \ [ \ m \ ] \ \{ \ g_e \ \} \ c_r \ [ \ SEP \ ] \ [ \ m \ ] \text{ is}
\end{equation*}
where $c_l, c_r$ denote context, $m$ the target mention, and $g_e$ its semantic group. The token $[SEP]$ and cue phrase “$m$ is” prompt the model to generate output sequence $y$ defined as:
\begin{equation*}
y = e_m
\end{equation*}
where $e_m$ is the concept corresponding to mention $m$. We maximize the conditional likelihood:
\begin{equation*}
p_\theta(y|x) = \prod_{i=1}^{N_y} p_\theta(y_i|y_{<i},x)
\end{equation*}
where $N_y$ is the number of tokens in the output, and $y_i$ is the $i$-th token.

\subsection{Adaptive Concept Representation for Training}
\begin{figure}[t]
    \centering
    \includegraphics[width=0.9\linewidth]{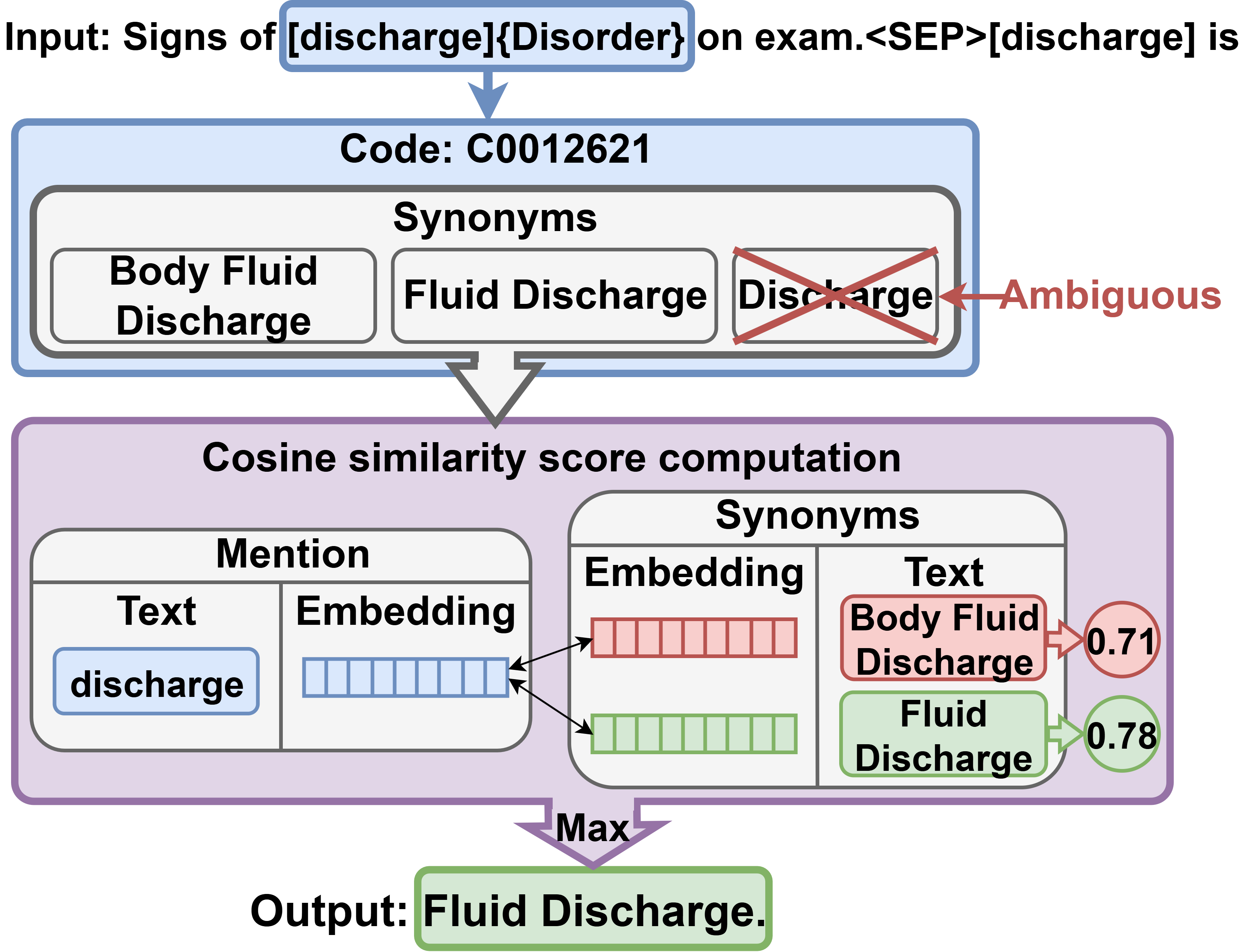}
    \caption{Example of the disambiguation process used to select a natural representation $e_m$ = ``Fluid Discharge'' for a concept $e$ = ``C0012621'' and a mention $m$ = ``discharge''.}
    \label{fig:ambiguity}
\end{figure}

A key challenge during training is representing the concept $e_m$ in natural language for the target output. We employ an adaptive method that selects the most appropriate synonym of a concept based on the input mention. As shown in Figure~\ref{fig:ambiguity}, the method proceeds in two steps. First, we preprocess each concept’s list of synonyms by removing those that are ambiguous with other concepts in the same semantic group. For example, the term “discharge” is removed from the \textit{Disorder} group because it can refer to multiple clinical symptoms. Then, for each remaining unambiguous synonym $s_e$, we compute the cosine similarity between the embeddings of the mention $m$ and $s_e$. The concept representation $e_m$ is defined as the synonym with the highest similarity score. 
$$
e_m = \arg \max_{s_e \in \mathcal{S}_e} \ \frac{\mathbf{v}_m \cdot \mathbf{v}_{s_e}}{\|\mathbf{v}_m\| \|\mathbf{v}_{s_e}\|}
$$
where $\mathbf{v}_m$ and $\mathbf{v}_{s_e}$ are the embedding vectors of the mention $m$ and synonym $s_e$, respectively. Importantly, the embeddings can be derived from any model, in our experiments, we compare different embedding models to assess their impact on linking performance.

\subsection{Synthetic Data Generation}
 To improve concept coverage while maintaining contextual realism, we generate synthetic training data using a structured prompt-based approach. For a given concept $e$, our method generates contextual sentences that capture diverse, natural ways of expressing $e$. As illustrated in Figure~\ref{fig:overview}, the prompt contains three components: (i) a task description that decomposes generation into two steps, first generating mentions of the concept and then generating contextualized sentences for each mention. (ii) Random contextual examples from the human-annotated training set. (iii) The title, semantic group, semantic type, definitions, and synonyms of the target concept from the KB. The prompt is written in the same language as the target dataset and provided to an LLM, which produces the final contextualized sentences. 

\subsection{Training Data Composition}

The synthetic examples generated by SynCABEL are combined with human-annotated training data to fine-tune the entity linking model (Figure~\ref{fig:overview}). To preserve the characteristics of clinical reports while benefiting from the increased coverage of synthetic data, we jointly train on both sources and upsample human-annotated examples such that they constitute half of the training instances.

\subsection{Guided Inference}

\begin{figure*}
    \centering
    \includegraphics[width=\textwidth]{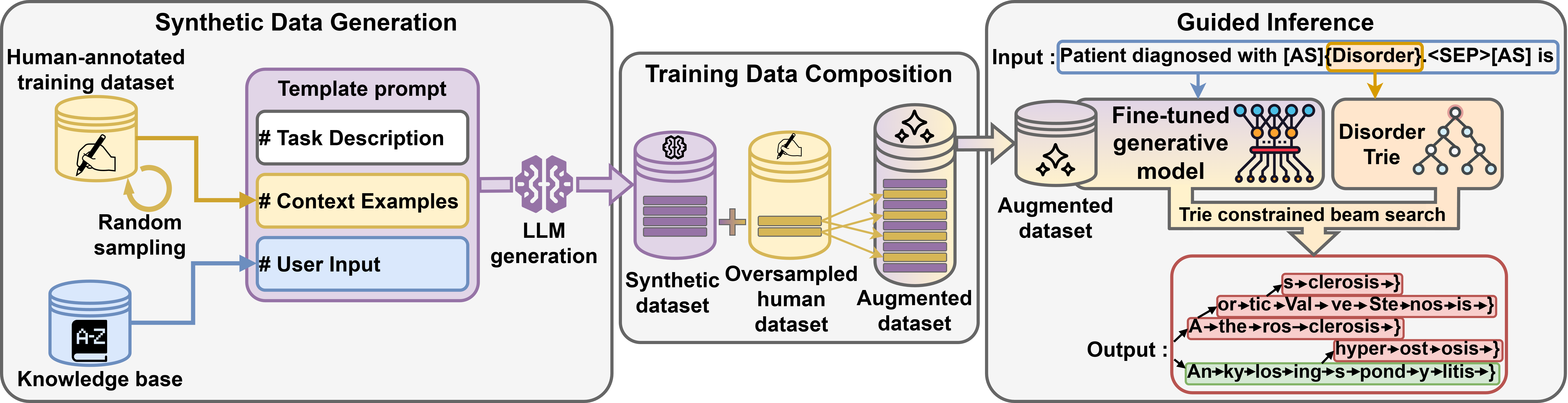}
    \caption{Overview of SynCABEL. LLM: large language model}
    \label{fig:overview}
\end{figure*}

Greedy decoding often yields invalid entities absent from the KB. To address this, we adopt the guided inference mechanism from~\citet{cao_autoregressive_2021}, illustrated in Figure~\ref{fig:overview}. First, the KB vocabulary is filtered by the target entity's semantic group (e.g., \textit{Disorders}) and structured as a trie, where each path maps a synonym to a unique concept. Then, the model dynamically restricts token selection at each step to valid branches. For instance, after generating the prefix ``A'', the trie permits only valid continuations like ``the'' (for \textit{Atherosclerosis}) or ``ortic'' (for \textit{Aortic}).

\subsection{LLM-as-a-judge Evaluation}
\begin{figure}
    \centering
    \includegraphics[width=0.9\linewidth]{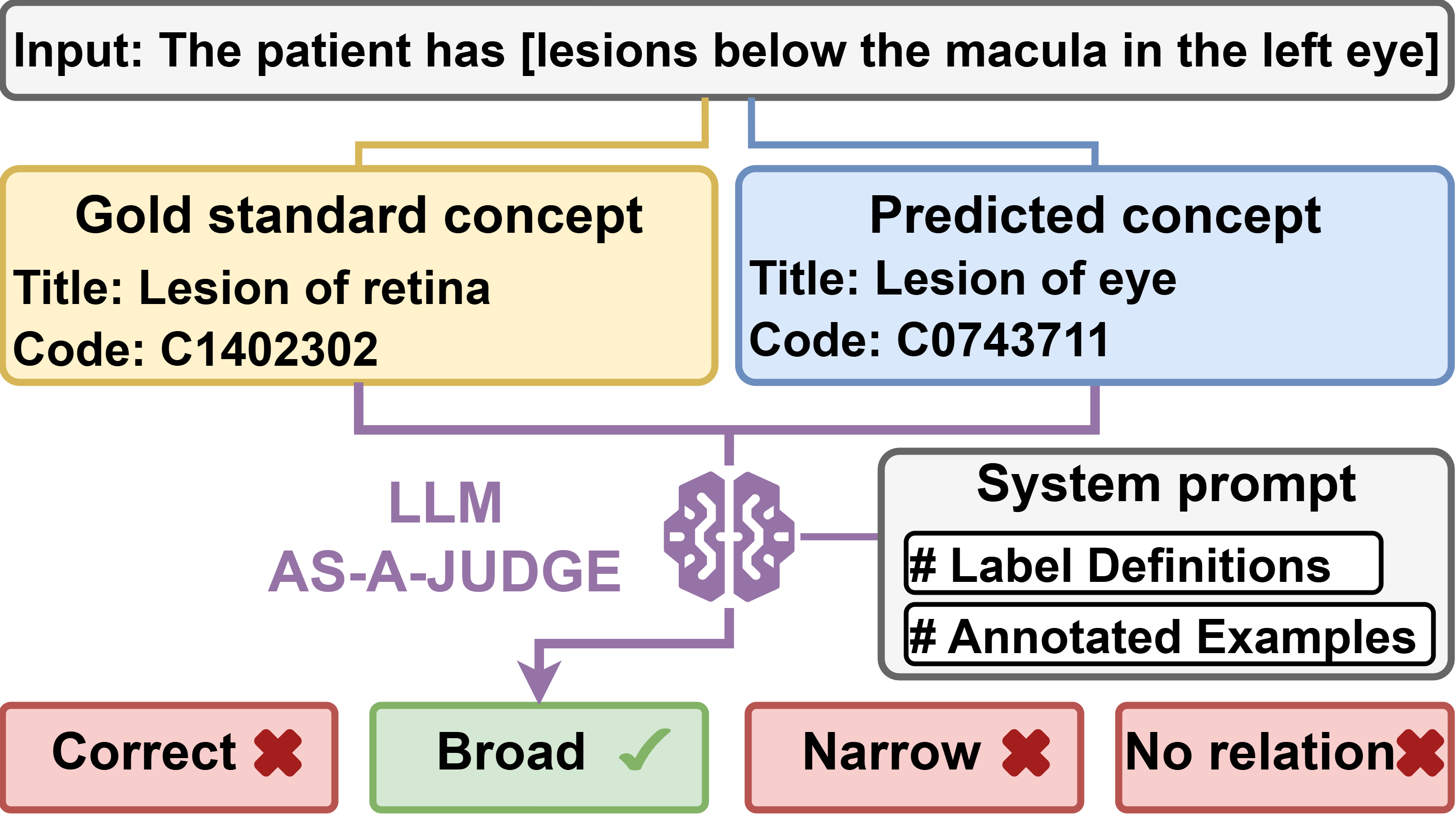}
    \caption{LLM-as-a-judge evaluation of predicted vs.\ gold concepts using four clinical-relation classes: Correct, Broad, Narrow, and No relation.}
    \label{fig:llm_as_a_judge}
\end{figure}

Standard entity linking evaluation relies on exact code matching and therefore may fail to capture the true clinical relationship between predicted and gold concepts. In practice, a prediction can be clinically correct despite a code mismatch due to ontology redundancy (e.g., multiple codes for the same concept) or annotation noise, where the prediction better reflects the mention than the human label. To capture these nuances, we introduce a four-class scheme (Figure~\ref{fig:llm_as_a_judge}): \textbf{(i) Correct}, clinically appropriate even without exact code match; \textbf{(ii) Broad}, relevant but overly general; \textbf{(iii) Narrow}, relevant but overly specific; and \textbf{(iv) No relation}, clinically unrelated. Labels are assigned automatically by an LLM using precise class definitions and five clinician-annotated examples.

\section{Experiments}

\subsection{Datasets}

\paragraph{Human-Annotated Datasets.}
We rely on three human-annotated corpora for our experiments:
\begin{itemize}
    \item \textbf{MedMentions-ST21pv (MM-ST21pv)}~\cite{mohan_medmentions_2018}, a curated subset of the MedMentions corpus, consisting of 4,392 English PubMed abstracts. MedMentions is currently the largest publicly available human-annotated dataset for biomedical entity linking.
    \item \textbf{QUAERO French Medical Corpus (QUAERO)} \cite{neveol_quaero_2014}, the largest French dataset for biomedical entity linking, composed of two subsets: QUAERO-MEDLINE, derived from MEDLINE titles, and QUAERO-EMEA, extracted from documents published by the European Medicines Agency.
    \item \textbf{Spanish Clinical Case Corpus (SPACCC)}~\cite{miranda-escalada_overview_2022,lima-lopez_overview_2023,lima-lopez_overview_2023-1}, a manually annotated collection of clinical case reports derived from open-access Spanish medical publications. The corpus contains 1,000 clinical cases, with disease, procedure, and symptom mentions annotated using the SNOMED CT ontology.
\end{itemize}
MM-ST21pv and QUAERO were selected because they are the largest human-annotated biomedical entity linking datasets available in their respective languages and because they cover nearly all UMLS semantic groups. SPACCC was chosen as the largest Spanish BEL dataset and, more importantly, because it consists of clinical reports, making it particularly valuable for assessing the potential applicability of our approach in clinical settings. Table~\ref{tab:dataset-summary} summarizes the main characteristics of these datasets.

\begin{table}
\centering
\setlength{\tabcolsep}{3.5pt}
\begin{tabular}{@{}p{2.1cm}
                >{\centering\arraybackslash}m{2.2cm}
                >{\centering\arraybackslash}m{1.7cm}
                >{\centering\arraybackslash}m{1.6cm}@{}}
\toprule
 & \textbf{MM-ST21PV}
 & \textbf{QUAERO}
 & \textbf{SPACCC} \\
\midrule
\# Docs & 4,392 & 2,422 & 1,000 \\
\# Mentions & 203,282 & 16,185 & 27,799 \\
\# Concepts & 25,419 & 5,045 & 9,254 \\
\# Sem. Groups & 14 & 10 & 3 \\
Language & English & French & Spanish \\
Clinical cases & \xmark & \xmark & \cmark \\
\bottomrule
\end{tabular}
\caption{Summary statistics of the human-annotated datasets. MM-ST21pv: MedMentions-ST21pv; SPACCC: Spanish Clinical Case Corpus}
\label{tab:dataset-summary}
\end{table}

\paragraph{Synthetic Datasets.}

We constructed three synthetic datasets (SynthMM, \mbox{SynthQUAERO}, and SynthSPACCC) by using \mbox{Llama-3-70B}~\cite{grattafiori_llama_2024} to generate three contextualized training examples per concept, prompted with five random annotated examples from the target training set. For instance, SynthMM includes the concept \textit{Apophysitis} (CUI C0264110) in the sentence: ``The adolescent athlete presented with knee pain and swelling, diagnosed with [apophysitis] of the tibial tubercle, highlighting the importance of proper training and warm-up routines.'' While applicable to all candidate concepts, SynthMM and SynthQUAERO were restricted to UMLS concepts with definitions (6.5\% and 4.5\% of the KB, respectively) for computational feasibility. Table~\ref{tab:synthetic-summary} summarizes the generated datasets. The full datasets, examples, and prompts are available on \href{https://hf.co/datasets/Aremaki/SynCABEL}{HuggingFace}.

\begin{table}
\centering
\begin{tabular}{@{}p{1.6cm}
                >{\centering\arraybackslash}m{1.3cm}
                >{\centering\arraybackslash}m{2.1cm}
                >{\centering\arraybackslash}m{2.3cm}@{}}
\toprule
\textbf{Dataset} & \textbf{SynthMM} & \textbf{SynthQUAERO}& \textbf{SynthSPACCC} \\
\midrule
\# Mentions                   & 460,878   & 396,914  & 1,813,463 \\
\# Concepts & 153,374   & 132,089 & 276,649  \\
\bottomrule
\end{tabular}
\caption{Summary of synthetic datasets.}
\label{tab:synthetic-summary}
\end{table}

\subsection{Implementation details}

\paragraph{Hardware and Software.}
All experiments were conducted on a single NVIDIA H100 GPU (80GB). For reproducibility, full implementation details, hyperparameters, and code are provided on \href{https://github.com/Aremaki/SynCABEL}{GitHub}.

\paragraph{Knowledge Bases.}
We used three KBs aligned with each dataset's annotation guidelines. For MM‑ST21pv, we used UMLS 2017AA restricted to the ST21‑pv scheme (21 semantic types and synonyms from 18 source ontologies). For QUAERO, we used UMLS 2014AA filtered to 10 semantic groups. For SPACCC, we relied on gazetteers derived from relevant branches of Spanish SNOMED-CT (July 31, 2021 release); since SPACCC comprises three subsets (diseases, symptoms, procedures), these categories served as semantic groups.

\begin{table*}[ht]
\centering
\begin{tabular}{lcccc|c}
\toprule
\textbf{Model} &
\shortstack{\textbf{MM-}\\\textbf{ST21PV}\\(english)} &
\shortstack{\textbf{QUAERO-}\\\textbf{MEDLINE}\\(french)} &
\shortstack{\textbf{QUAERO-}\\\textbf{EMEA}\\(french)} &
\shortstack{\textbf{SPACCC}\\(spanish)} &
\textbf{Average} \\
\midrule
\multicolumn{6}{l}{\textsc{Rule-based (Unsupervised)}} \\
\midrule
SciSpacy~\cite{neumann_scispacy_2019} & 53.8 & 40.5 & 37.1 & 13.2 & 36.2 \\
\midrule
\multicolumn{6}{l}{\textsc{Context-free Bi-encoder (Self-supervised)}} \\
\midrule
SapBERT~\cite{liu_self-alignment_2021} & 51.1 & 50.6 & 49.8 & 33.9 & 46.4 \\
CODER-all~\cite{yuan_coder_2022} & 56.6 & 58.7 & 58.1 & 43.7 & 54.3 \\
SapBERT-all~\cite{liu_learning_2021} & 64.6 & 74.7 & 67.9 & 47.9 & 63.8 \\
\midrule
\multicolumn{6}{l}{\textsc{Contextualized Bi-encoder (supervised)}} \\
\midrule
ArboEL~\cite{agarwal_entity_2022} & \underline{74.5} & 70.9 & 62.8 & 49.0 & 64.2 \\
\midrule
\multicolumn{6}{l}{\textsc{Generative encoder-decoder (supervised)}} \\
\midrule
mBART-large~\cite{tang_multilingual_2020} & 65.5 & 61.5 & 58.6 & 57.7 & 60.8 \\
+ Guided inference~\cite{cao_autoregressive_2021} & 70.0 & 72.8 & 71.1 & 61.8 & 68.9 \\
\rowcolor{blue!10}
\textbf{+ SynCABEL (Our method)} & 71.5 & 77.1 & \underline{75.3} & 64.0 & 72.0 \\
\midrule
\multicolumn{6}{l}{\textsc{Generative decoder-only (supervised)}} \\
\midrule
Llama-3-8B~\cite{grattafiori_llama_2024} & 69.0 & 66.4 & 65.5 & 59.9 & 65.2 \\
+ Guided inference~\cite{cao_autoregressive_2021} & 74.4 & \underline{77.5} & 72.9 & \underline{64.2} & \underline{72.3} \\
\rowcolor{blue!10}
\textbf{+ SynCABEL (Our method)} & \textbf{75.4} & \textbf{79.7} & \textbf{79.0} & \textbf{67.0} & \textbf{75.3} \\
\bottomrule
\end{tabular}
\caption{
Entity linking performance (Recall@1) on biomedical benchmarks. The \colorbox{blue!10}{highlighted} rows show our contribution, where the base model is trained on an augmented dataset. The best results are shown in \textbf{bold}, the second-best results are \underline{underlined}, and the ``Average'' column reports the mean score across the four benchmarks.
}

\label{tab:benchmark-results}
\end{table*}

\subsection{Results}

\subsubsection{Benchmark Performance}

We reproduced several state-of-the-art BEL systems from different paradigms: the rule-based \textbf{SciSpacy}~\cite{neumann_scispacy_2019}; the context-free bi-encoders \textbf{SapBERT}~\cite{liu_self-alignment_2021}, \textbf{SapBERT-all}~\cite{liu_learning_2021}, and \mbox{\textbf{CODER-all}}~\cite{yuan_coder_2022}; the contextualized bi-encoder \mbox{\textbf{ArboEL}}~\cite{agarwal_entity_2022}; the generative encoder-decoder \mbox{\textbf{mBART-large}}~\cite{tang_multilingual_2020} and decoder-only \textbf{Llama-3-8B}~\cite{grattafiori_llama_2024}. For generative models (mBART-large and Llama-3-8B), we evaluated three configurations: (i) supervised fine-tuning on human-annotated data; (ii) adding guided inference adapted from GENRE~\cite{cao_autoregressive_2021}; and (iii) our proposed SynCABEL approach, which augments the training data with synthetic examples while retaining guided inference. For fair comparison, all baselines were constrained to the same KBs, and for each mention, only concepts from the matching semantic group were considered as candidates. In practice, this meant that for a given input, every baseline was provided with the same candidate set. This constraint could not be applied to SciSpacy, as its rule-based approach uses the entire KB and cannot be modified.

Table~\ref{tab:benchmark-results} presents entity linking performance (Recall@1) across four biomedical benchmarks: MM-ST21pv, QUAERO-MEDLINE, QUAERO-EMEA, and SPACCC. Models are grouped by architecture paradigm. Our SynCABEL framework with \mbox{Llama-3-8B} achieves the highest scores on all four benchmarks: 75.4 on MM-ST21pv, 79.7 on QUAERO-MEDLINE, 79.0 on QUAERO-EMEA, and 67.0 on SPACCC. The mBART-large variant of SynCABEL also shows consistent improvements over its baseline across all datasets.

\subsubsection{Adaptive Concept Representation for Training}

We evaluate our adaptive concept representation strategy against a static baseline using concept preferred titles. We consider two adaptive representations: character-level 3-gram TF-IDF~\cite{neumann_scispacy_2019}, following its effective use in GenBioEL~\cite{yuan_generative_2022} and CODER-all embeddings~\cite{yuan_coder_2022}. Experiments are conducted on UMLS-based datasets (MM-ST21pv, QUAERO-MEDLINE and QUAERO-EMEA) using mBART-large, with training limited to the original data (no augmentation) and guided decoding at inference.
\begin{table}
\centering
\small
\setlength{\tabcolsep}{3.5pt}
\begin{tabular}{@{}p{1.7cm}
                >{\centering\arraybackslash}m{1.5cm}
                >{\centering\arraybackslash}m{1.8cm}
                >{\centering\arraybackslash}m{1.8cm}@{}}
\toprule
 & \shortstack{\textbf{MM-}\\\textbf{ST21PV}} 
 & \shortstack{\textbf{QUAERO-}\\\textbf{MEDLINE}} 
 & \shortstack{\textbf{QUAERO-}\\\textbf{EMEA}} \\
\midrule
Title (Static) & 61.7 & 53.7 & 50.6 \\
CODER-all & 68.4 & 72.5 & 66.0 \\
\rowcolor{blue!10}
TF-IDF & \textbf{70.0} & \textbf{72.8} & \textbf{71.1} \\
\bottomrule
\end{tabular}
\caption{Recall@1 scores for different concept representation methods. mBART-large was supervised only on the human-annotated dataset (i.e., without data augmentation), and inference was performed using the guided inference method. The \colorbox{blue!10}{highlighted} row shows the chosen representation method.}
\label{tab:concept_representation}
\end{table}
\\ As shown in Table~\ref{tab:concept_representation}, the TF-IDF-based representation consistently achieves the highest Recall@1 across all datasets, outperforming both the static baseline and CODER-all. We therefore adopt TF-IDF-based adaptive representations for all other experiments.

\subsubsection{Training Data Composition}

We evaluate three training data composition strategies when combining human-annotated and synthetic data: \textbf{(i) Synthetic Pretrain}, which uses a two-stage training procedure, first training on synthetic data and then fine-tuning on human data, without resampling;
\textbf{(ii) Combined}, which performs a single-stage training on the union of human and synthetic datasets as-is, without resampling;
\textbf{(iii) Interleaved}, which performs a single-stage training on merged data while oversampling human samples so that each training step alternates between a human and a synthetic instance. These experiments use \mbox{Llama-3-8B} with guided inference.
\begin{table}
\centering
\setlength{\tabcolsep}{3.5pt}
\begin{tabular}{lcccc}
\toprule
& \shortstack{\textbf{MM-}\\\textbf{ST21pv}} 
& \shortstack{\textbf{QUAERO-}\\\textbf{MEDLINE}} 
& \shortstack{\textbf{QUAERO-}\\\textbf{EMEA}} 
& \textbf{SPACCC} \\
\midrule
SPT
& 74.1 
& 78.0 
& 77.1 
& 64.0 \\
COMB
& 75.4 
& 79.4 
& 77.0 
& 64.3 \\
\rowcolor{blue!10}
INT 
& \textbf{75.4} 
& \textbf{79.7} 
& \textbf{79.0} 
& \textbf{67.0} \\
\bottomrule
\end{tabular}
\caption{Recall@1 scores for different training data composition strategies. The \colorbox{blue!10}{highlighted} row shows the chosen strategy.}
\label{tab:training_composition}
\end{table}
\\ Recall@1 results in Table~\ref{tab:training_composition} show that interleaved oversampling consistently outperforms other strategies across all datasets, highlighting the importance of preserving a strong human signal while leveraging synthetic diversity. We therefore adopt this strategy in other experiments.

\subsubsection{LLM-as-a-Judge Evaluation}

To complement exact code matching, we employed GPT-5.2\footnote{Developed by OpenAI and accessed via their API.} as an automated judge to assess clinical validity on SPACCC, the only dataset containing full clinical reports. We evaluated all cases where the prediction of our best-performing model (\mbox{Llama-3-8B}) differed from the gold standard.\footnote{We also explored the KB-based architecture to explicitly detect broader/narrower relations,
but it did not yield reliable results, likely due to the complexity and brittleness of these KBs.}

A clinician labeled 150 mismatches across Disorders, Symptoms, and Procedures. GPT-5.2 achieved 66.2\% agreement, with 79\% precision for the `Correct' label. Crucially, errors are conservative: only 12.4\% of judge predictions overstated the human label (e.g., Narrow for No relation). This tendency to under-predict relevance implies the judge acts as a lower-bound estimator of validity.

Table~\ref{tab:llm_results} compares standard exact matching against the LLM-as-a-judge evaluation metric. While SynCABEL yields a 2.8\% gain in exact matching, it achieves a larger 3.7\% improvement in clinically valid predictions (rising from 68.9\% to 72.6\%). Given the judge's proven conservative bias, it shows \mbox{SynCABEL} helps the model capture correct semantics even when it misses the exact target code.

\begin{table}
\centering
\begin{tabular}{lll}
\toprule
\textbf{Label} & \textbf{w/o SynCABEL} & \textbf{w/ SynCABEL} \\
\midrule
Exact Matching                    & 64.2 (62.6–65.8)  & 67.0  (65.5–68.6) \\
\midrule
Correct                    & \textbf{68.9 (67.4–70.4)}   & \textbf{72.6  (71.2–74.1)} \\
Broad                   & 13.0  (12.1–13.8)   & 11.0  (10.3–11.8) \\
Narrow                   & \ \ 4.3  (3.8–4.7)   & \ \ 5.9  (5.3–6.5)\\
No relation               & 13.8 (11.1–16.7) & 10.5 (7.6–13.2)  \\
\bottomrule
\end{tabular}
\caption{Llama-3-8B performance on SPACCC with and without SynCABEL augmentation. Results are reported as percentages with 95\% confidence intervals estimated via document-level empirical bootstrap, for exact code matching and each LLM-as-a-judge label.}
\label{tab:llm_results}
\end{table}

\section{Discussion}

\subsection{Annotation Scarcity}

To assess synthetic data impact, we stratify performance by whether concepts appeared in training. As shown in Table~\ref{tab:seen-unseen-table}, SynCABEL consistently improves unseen concepts performance across benchmarks while preserving seen results, with notable gains on SPACCC and QUAERO-EMEA (+9.4 and +9.9 points). However, unseen performance remains below seen (e.g., 30.2 vs 83.0 on SPACCC), showing that SynCABEL reduces but does not fully close the annotation scarcity gap. For UMLS-based datasets (QUAERO, MM), generation was restricted to concepts with definitions, leaving some ``unseen'' test concepts absent from augmented data, a filter that limits gains compared to SPACCC, suggesting that extending generation to all concepts could further bridge this gap.

\begin{table}[h]
\centering
\begin{tabular}{lll}
\toprule
\textbf{Subset} & \textbf{w/o SynCABEL} & \textbf{w/ SynCABEL} \\
\midrule
\multicolumn{3}{l}{\textsc{MedMentions-ST21pv}} \\
\midrule
Overall & 74.4 (73.6–75.4) & 75.4 (74.5–76.3) \\
Seen Concepts & 81.8 (81.0–82.6) & 82.2 (81.4–83.0) \\
Unseen Concepts & 46.0 (43.9–48.2) & 49.0 (46.7–51.3) \\
\midrule
\multicolumn{3}{l}{\textsc{QUAERO-MEDLINE}} \\
\midrule
Overall & 77.5 (75.9–79.1) & 79.7 (78.0–81.2) \\
Seen Concepts & 90.5 (89.1–91.8) & 89.7 (88.3–91.0) \\
Unseen Concepts & 58.7 (55.8–61.6) & 65.1 (62.1–68.1) \\
\midrule
\multicolumn{3}{l}{\textsc{QUAERO-EMEA}} \\
\midrule
Overall & 72.9 (68.3–77.3) & 79.0 (75.7–82.2) \\
Seen Concepts & 89.0 (86.1–91.6) & 92.0 (89.6–94.3) \\
Unseen Concepts & 52.5 (46.1–59.3) & 62.4 (56.4–68.7) \\
\midrule
\multicolumn{3}{l}{\textsc{SPACCC}} \\
\midrule
Overall & 64.2 (62.6–65.8) & 67.0 (65.5–68.6) \\
Seen Concepts & 83.0 (81.8–84.2) & 83.0 (81.8–84.2) \\
Unseen Concepts & 20.8 (19.2–22.6) & 30.2 (28.1–32.4) \\
\bottomrule
\end{tabular}
\caption{Exact-match Recall@1 performance on three biomedical benchmarks comparing two variants of Llama-3-8B: one trained only on the corresponding human-annotated dataset with guided inference, and one further enhanced with SynCABEL. Results are reported overall and for subsets defined by seen/unseen concepts. 95\% confidence intervals were estimated via the empirical bootstrap method at the document level.}
\label{tab:seen-unseen-table}
\end{table}

\subsection{Reducing Annotation Effort}

We evaluate data efficiency by varying the fraction of human-annotated training data (0--100\%), with subsets created by randomly sampling documents, and comparing SynCABEL augmentation to the human-annotated only baseline.
\begin{figure}
    \centering
    \includegraphics[width=\linewidth]{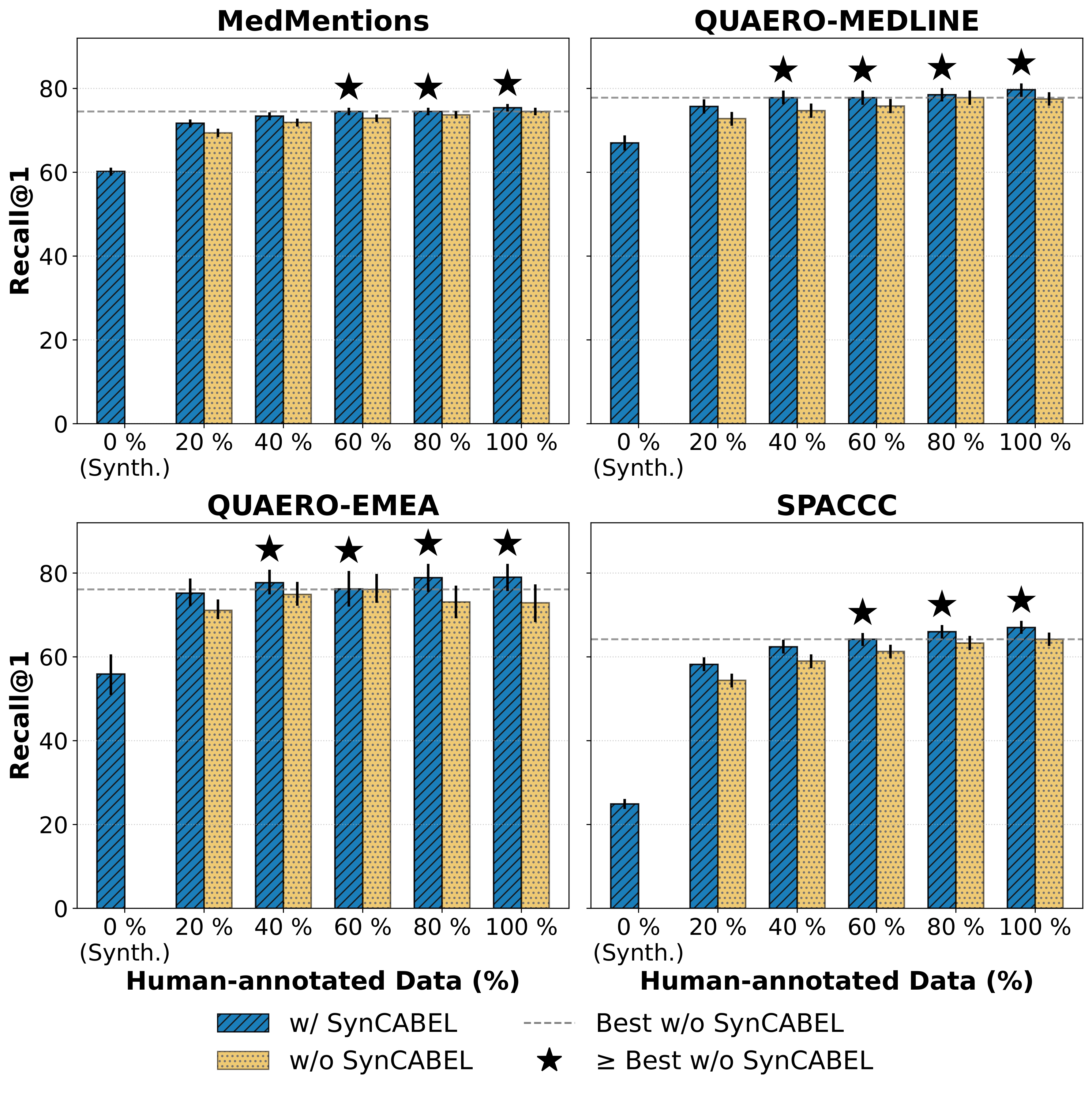}
    \caption{Data Efficiency Analysis. Exact-match Recall@1 performance as a function of human training data availability. The dashed line marks the performance ceiling of the standard baseline trained on human-annotated data. Stars ($\star$) indicate where the SynCABEL-augmented model (blue) matches or surpasses this ceiling. 0\% implies training on synthetic data only. Error bars denote 95\% bootstrap CIs.}
    \label{fig:results_recall_comparison}
\end{figure}
\\ Figure~\ref{fig:results_recall_comparison} demonstrates that SynCABEL substantially reduces annotation needs: it reaches full-data performance using 60\% of human annotations on MM-ST21pv and SPACCC, and 40\% on QUAERO. The gains are largest in low-resource regimes and diminish as more human data becomes available. This trend is clear on SPACCC and MM-ST21pv, though it is less observable on QUAERO due to the larger confidence intervals inherent to its smaller size. Finally, models trained solely on synthetic data significantly underperform fully supervised baselines, confirming that human annotations remain essential for optimal results on human-annotated data.

\subsection{Real-World Applicability}
BEL plays a central role in enabling the secondary use of electronic health records. By transforming unstructured free-text into structured concepts, it supports medical research, clinical decision, and downstream tasks such as automated clinical coding, cohort identification, and patient recruitment for clinical trials~\cite{remaki_improving_2025,french_overview_2023}. To assess the viability of our models for such real-world deployment, we evaluate two  dimensions: the utility of confidence scores for ensuring high-precision predictions, and the computational efficiency required for scale.

\paragraph{Prediction Confidence.}
For each prediction, our model outputs a confidence score, computed as the mean token probability of the generated sequence. This score enables confidence-based filtering to trade precision for recall depending on application needs. On SPACCC, we empirically set a confidence threshold of 0.9 and evaluated performance on predictions exceeding this threshold.

It yields a precision of 80.8\% (79.6–81.9), recall of 61.9\% (60.2–63.5), and F1-score of 70.1\% (68.5–71.6). These results illustrate how applying a strict confidence threshold can substantially boost precision for high-stakes applications while retaining reasonable recall.

\paragraph{Inference Efficiency and Memory Footprint.}
We evaluate real-world deployability based on throughput and memory usage (Table~\ref{tab:efficiency}). While SapBERT achieves the highest speed, it requires a massive 20.1~GB candidate index. \mbox{ArboEL} uses a smaller BERT encoder, reducing model size (1.2~GB) and candidate memory (7.1~GB), but throughput is lower (38.9~mentions/s) due to the cross‑encoder re‑ranking step required at inference. In contrast, generative models rely on a compact candidate trie (5.4~GB), independent of model size. The mBART configuration processes 51.0 mentions/s with a small model footprint (2.3~GB), whereas \mbox{Llama-3-8B} requires substantially more GPU memory (28.6~GB) and achieves lower throughput (19.1 mentions/s).

\begin{table}
\centering
\begin{tabular}{lccc}
\toprule
 & Model (GB) & Cand. (GB) & Speed (/s) \\
\midrule
SapBERT & 2.1 & 20.1 & \textbf{575.5} \\
ArboEL & \textbf{1.2} & 7.1 & 38.9 \\
mBART & 2.3 & \textbf{5.4} & 51.0 \\
Llama-3-8B & 28.6 & \textbf{5.4} & 19.1 \\
\bottomrule
\end{tabular}
\caption{Inference speed and memory footprint. Model: Model parameters size; Cand.: Candidate memory size.}
\label{tab:efficiency}
\end{table}

\section{Conclusion and Future Work}

We introduced SynCABEL, a framework addressing annotation scarcity in BEL by generating synthetic, contextualized training examples for all candidate concepts in a target KB. By systematically enriching concept representations with diverse usage contexts, SynCABEL achieves state-of-the-art performance on multilingual benchmarks, improves recall on unseen concepts, and increases clinical validity under our LLM-as-a-judge protocol. While human annotations remain the strongest form of supervision, our results show that competitive performance can be achieved with substantially less of this costly data, thereby maximizing the value of expert effort.

Future directions include extending the generation context beyond single sentences to better capture document-level evidence, expanding to additional languages and domains, and refining training with negative sampling strategies such as ANGEL~\cite{kim_learning_2025}. We also aim to improve synthetic data quality through prompt refinement and expert intrinsic evaluation, and to reduce generation cost through smaller LLMs and selective concept sampling.

\section*{Ethical Statement}

This work does not use private or identifiable patient records. Experiments rely on publicly available or licensed biomedical entity linking resources, together with synthetic data generated for research purposes. QUAERO was used under GFDL, MedMentions under CC0, and the UMLS Metathesaurus under its License Agreement, with acknowledgment of the National Library of Medicine. The released synthetic datasets are intended for research in biomedical entity linking and should not be used for clinical decision-making without appropriate validation.

Although synthetic data reduces privacy risks and dependence on costly expert annotation, it may still reflect biases, errors, or limitations of the source knowledge bases and generative models. We therefore encourage users to inspect the data before downstream use. Generating large-scale synthetic data also incurs computational and environmental costs; future work should balance data volume, model performance, and energy use. The authors used GitHub Copilot for code completion; all code was reviewed by the authors, who take full responsibility for it.

\section*{Acknowledgments}

This work was performed using HPC resources from GENCI-IDRIS (Grant 2025-AD010316788). The authors thank the NLP for Biomedical Information Analysis team at the Barcelona Supercomputing Center for their collaboration. This work was funded by the Région Île-de-France through the AI4IDF doctoral scholarship and by government funding managed by the French National Research Agency (ANR) under the France 2030 program, grant ANR-23-IACL-0007.

\bibliographystyle{named}
\bibliography{ijcai26}

\end{document}